\documentclass{article}
\usepackage[preprint]{neurips_2020}
\usepackage{amsmath}
\usepackage[utf8]{inputenc} 
\usepackage[T1]{fontenc}    
\usepackage{hyperref}       
\usepackage{url}            
\usepackage{booktabs}       
\usepackage{amsfonts}       
\usepackage{nicefrac}       
\usepackage{microtype}      
\usepackage{graphicx}

\title{Causal inference using deep neural networks}

\author{%
  Ye Yuan\thanks{Contributed equally} \\
  Machine Learning Department\\
  School of Computer Science\\
  Carnegie Mellon University\\
  Pittsburgh, Pennsylvania, 15213\\
  \And
  Xueying Ding*\\
  Machine Learning Department\\
  School of Computer Science\\
  Carnegie Mellon University\\
  Pittsburgh, Pennsylvania, 15213\\
  \And
  Ziv Bar-Joseph\\
  Computational Biology Department and Machine Learning Department\\
  School of Computer Science\\
  Carnegie Mellon University\\
  Pittsburgh, Pennsylvania, 15213\\
  \texttt{zivbj@andrew.cmu.edu}
}

\begin{document}

\maketitle

\begin{abstract}
Causal inference from observation data is a core problem in many scientific fields. Here we present a general supervised deep learning framework that infers causal interactions by transforming the input vectors to an image-like representation for every pair of inputs. Given a training dataset we first construct a normalized empirical probability density distribution ($NEPDF$) matrix. We then train a convolutional neural network (CNN) on $NEPDF$s for causality predictions. We tested the method on several different simulated and real world data and compared it to prior methods for causal inference. As we show, the method is general, can efficiently handle very large datasets and improves upon prior methods. 
\end{abstract}

\section{Introduction}

An important problem in many scientific areas is the inference of causal interactions between two variables from observational data. Examples include causal inference between proteins in biological studies ~\cite{yuan2019deep,marbach2012wisdom}, studies related to climate and global warming ~\cite{runge2019inferring} and studies of economic implications of various potential interventions~\cite{granger1969investigating}. Given its central role, several supervised and unsupervised methods have been developed for predicting such causal interactions. Unsupervised methods utilize various correlation metrics \cite{fonollosa2019conditional}, regression analysis ~\cite{granger1969investigating} or attempt to learn joint probability distributions using directed graphical models  ~\cite{koller2009probabilistic,spirtes2000causation}. Supervised methods often start with a few labeled samples and then use various classifiers or their modifications to predict causal interactions in test data \cite{guyoncause,lopez2015towards,mooij2016distinguishing}. While some recent work attempted to use deep learning techniques for causal inference \cite{goudet2018learning,kalainathan2018structural}, in many cases the way the observational data is obtained and constructed does not preserve any local information. Consider, for example, weather related data from several locations. For each location we may have several different values (avg temp, wind, precipitation etc.) but it is not clear in what order they should be provided for the network to fully utilize dependency between values that are in close proximity in the input space. A similar case occurs when genes in one cell are profiled. In such cases, the ordering of the input vectors is often arbitrary.

Here we extend regression methods for causality inference by enabling the use of deep neural networks (NN) for causal inference. For this, we extend a method first applied to inferring interactions in gene expression data \cite{yuan2019deep}. In contrast to gene expression data, where the main question is undirected interaction prediction, causality inference is non symmetric.  To infer causal interactions we reformat the input and obtain an image-like representation for the data. For this, we first convert the joint observations for each pair of samples to a normalized discrete probability matrix, where each entry represents the joint density of a set of observations. This preserves 
the local relationships between the values observed for each of the samples. Next, a CNN is used to classify based on these distributions. An overview of the method is illustrated in Figure~\ref{Figure 2}.

We tested our method on a series of simulated network structures including “V structure” ($X \rightarrow Z \leftarrow Y$), “chain structure” ($X \rightarrow Z \rightarrow Y$), and “reverse structure” ($X \leftarrow Y \rightarrow Z$), and also applied it to four real world datasets in various fields. In all cases our method improved on prior methods suggested for causal inference. 
\section{Related work}
\label{gen_inst}
The gold standards for causal inference are randomized controlled experiments in which $X$ is perturbed and its impact on $Y$ is studied \cite{varian2016causal,judea2000causality}. However, for many real world datasets such experiments are either impossible to perform (for example, manipulating genes within an individual) or very challenging \cite{khan2019genome}.

Another popular strategy for causal inference, which is often applied to time-series data, uses regression analysis. One of the most popular methods for such analysis is Granger causality. Granger causality inference learns two models $Y_t\left(X_{t'} \right)$ and $X_t\left(Y_{t'} \right)$ where $t' < t$ and compares between them to determine which better explains the other variable \cite{granger1969investigating,kim2011granger,finkle2018windowed}. Some methods extend Granger causality using supervised training. For example, SIGC  attempts to classify Granger causality results based on difference between conditional distributions observed in training data \cite{Chikahara2018CausalII}. Ganger causality, and other regression based causal inference methods can utilize either linear or non-linear regression \cite{kim2011granger, shimizu2011directlingam, peters2014causal,hoyer2009nonlinear}. However, the specific type of regression model (as opposed to the parameters) is a user defined issue and is not learned from the data.

A more global strategy attempts to learn joint efficient probability distributions based on directed edges. These methods mainly rely on variants of probabilistic graphical model (PGM) \cite{spirtes2000causation}. Unlike pairwise analysis methods, PGM methods can leverage information from other variables when determining the specific pairwise relationship between two variables. However, despite being directed, PGMs are not meant to provide causal relationship inference and in many cases edge directions in such networks do not imply causality \cite{nadkarni2001bayesian}.

Other methods focus primarily on causality inference for paired variables. One idea for inferring casual relationships between such variables is based on the independent additive noise assumption. Under such model we assume that if $X \rightarrow Y$, then the observed values will follow the equation: $Y = f(X) + noise $ \cite{shimizu2011directlingam,peters2014causal,inproceedings}. However, how to find a suitable $f$ and noise function form is a challenge for such methods  \cite{hoyer2009nonlinear}. Despite several of the methods that we will discuss as comparisons, an extension of such notion is the TiMINo model, which uses both additive noise and time information to infer causality from time-course data \cite{peters2013causal}. 

When labeled data is available, any one of several supervised methods can be used to perform such inference. One recent example is RCC which uses kernel mean embedding to construct features from sample pairs to train random forest classifiers \cite{lopez2015towards}. Another is based on neural networks for embedding samples in lower dimension before performing  causal inference \cite{lopez2017discovering}. See \cite{kalainathan2020causal} for a recent comprehensive review of causal inference methods. 

\section{Method}
\label{headings}
\subsection{Motivation}
\begin{figure}
  \centering
  \includegraphics[width=15.5 cm]{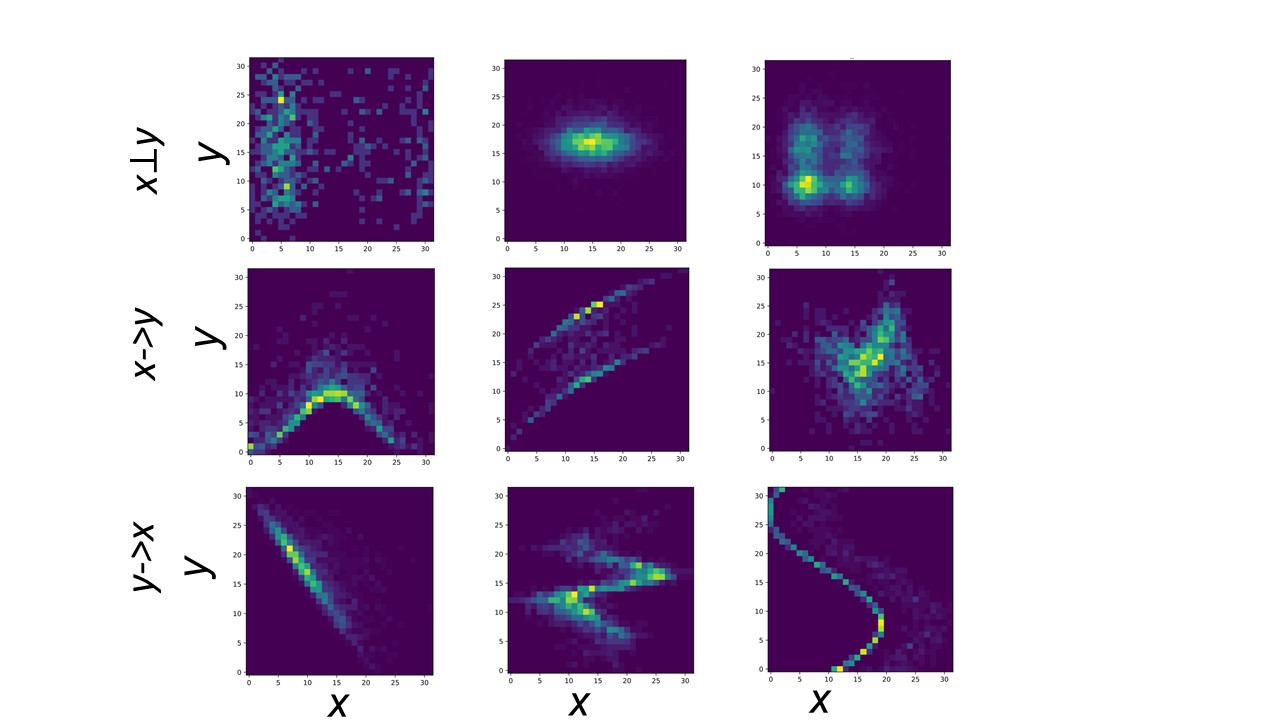}
  \caption{Causal samples ($X$ and $Y$ are independent in the upper panels, $X \rightarrow Y$ in the middle panels and $Y \rightarrow X$ in the bottom panels) from Kaggle 2013 NIPS challenge \cite{guyoncause}. Given paired samples for one variable pair ( $X, Y$) a normalized empirical probability distribution is constructed.}
   \label{Figure 1}
\end{figure}

\begin{figure}
  \centering
  \includegraphics[width=10.5 cm]{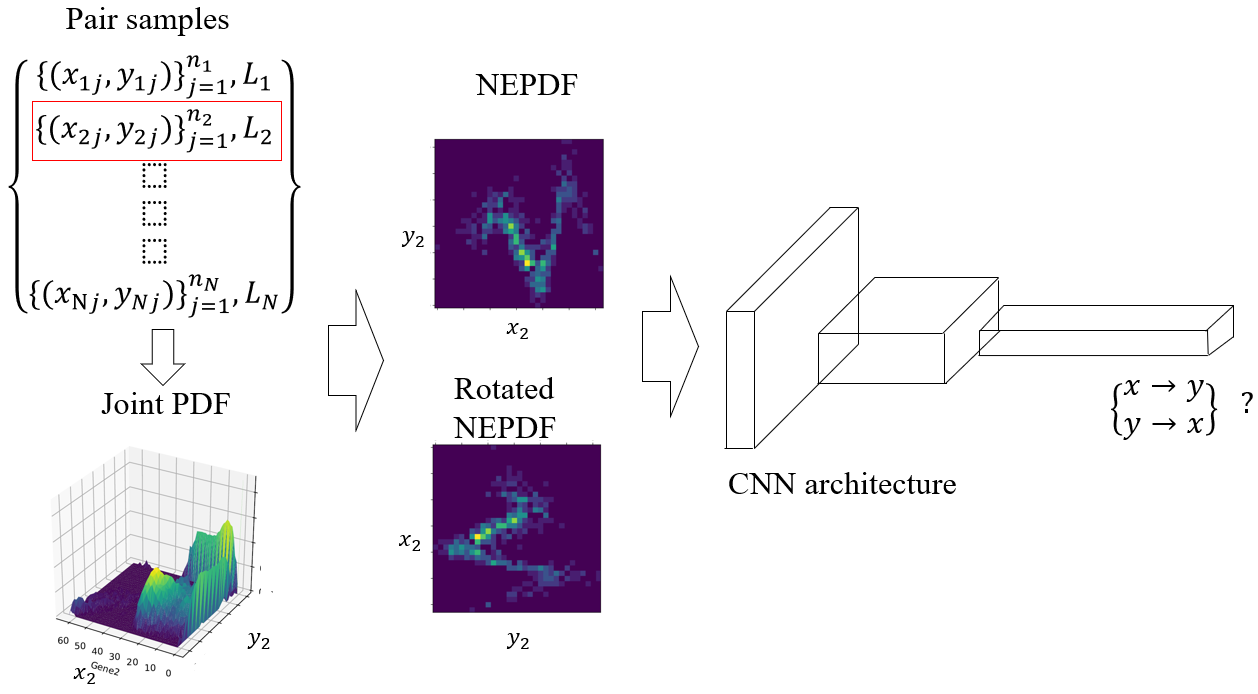}
  \caption{Method overview. Given paired samples for one variable pair a normalized empirical probability distribution function (NEPDF) is constructed. The NEPDF is used to train a CNN network for predicting causal interactions and relationships.}
   \label{Figure 2}
\end{figure}

For pair-wise causality inference tasks, most prior methods rely on specific assumptions about the type of noise (additive, multiplicative etc.) or the type of regression relationship expected (linear $f(X)$, nonlinear $f(X)$, Gaussian and other noise distributions \cite{kalainathan2020causal}). While such approaches work well for certain types of data, they may not be general enough to accommodate datasets that display mixtures of these noise or regression characteristics. For example, consider the causal relationships displayed in Figure~\ref{Figure 1} which are all part of the Kaggle causal-effect pair dataset \cite{guyoncause}. While some of these relationships are indeed linear, others are not and no specific regression model is likely to be able to accomodate all the different types of relationships observed in this data. 

To enable the use of general strategy for causal inference we developed a new method that replaces the regression component of current methods with a general CNN that can be used to learn both linear and non linear relationships depending on the input data. Specifically, consider the Granger's causality formulation \cite{granger1969investigating}, which learns a model for multiple time-series variables $X_1,X_2,...X_n$ with time-lag $p$ by setting : $X_j(t) =\sum_{i=1}^{n}\sum_{a=1}^{p}\beta_{ji,a}X_i(t-a) + \epsilon_j(t)$, where $\beta$ is the matrix of coefficients modeling the effect of time series $X_i$ on the time series target $X_j$. In this approach we need to set both the time lag and the types of coefficients we learn. Imn our model we replace this with a general CNN that can be trained to determine both an appropriate lag and the specific type of coefficients needed to accurately infer casuality as we discuss below. 

\subsection{Distribution classification}
Our first goal is to obtain a representation of the input data that can enable the use of convolutional neural networks for this task. Many causal problems use as input a collection samples. For example, in biology,  when attempting to learn gene-gene interactions we may have samples with the joint expression of these genes in thousands of cells. Such problems attempt to infer causality from an observation set and a small number of labeled pairs:
$\{\left(\{(x_{ij},y_{ij})\}_{j=1}^{n_i }),L_i \right)\}_{i=1}^N$,
where  $n_i$ is the number of observations for the $ith$ pair $(x_i,y_i)$, $N$ is the number of pairs and $L_i \in \{1,-1,0\}$ is the label for the causal interactions such that $L_i=1$ if $x_i \rightarrow y_i$ and
$L_i=-1 $ if $y_i \rightarrow x_i$. $L_i$ could also be 0 meaning the two variables are independent. One way to infer causal interactions from this data is by learning a mapping function: $
f:\mathbb{R}^{2\times n_i} \rightarrow \mathbb{L}
$ where $\mathbb{L}$ is label set, and $n_i$ is the number of observations for each pair.

We aim to develop an efficient method that can learn casual relationships without relying on a pre-determined statistical model. If we assume $(x_{ij},y_{ij}) \stackrel{\text{i.i.d}}{\sim} P_i$, where $P_i$ is the joint PDF of $(x_i,y_i)$, then the problem can be converted to distribution classification problem \cite{law2017bayesian,lopez2015towards,poczos2013distribution}: $
f:\mathcal{P} \rightarrow \mathbb{L}
$
where $\mathcal{P}$ is a joint PDF family and $P_i \in \mathcal{P}$. Note that in many cases the pairwise values are indeed independent between the samples and so this assumption holds for many real world datasets.

\subsection{NEPDF generation}

 Given a set of observations for two entities in our data $\left(\{(x_{ij},y_{ij})\}_{j=1}^{n_i }),L_i \right)$, the first step is to estimate the pair's joint PDF $\widetilde {P_i}$.  $\widetilde {P_i}$ is often estimated using a 2D kernel density estimation \cite{silverman1986density}:

\begin{equation}
\begin{split}
P(x_i,y_i) &= \lim_{h_{x_i} \rightarrow 0,h_{y_i} \rightarrow 0} \frac{1}{4h_{x_i}h_{y_i}}Pr[x_i-h_{x_i}<X_i<x_i+h_{x_i}, y_i-h_{y_i}<Y_i<y_i+h_{y_i}] \\
&= \lim_{h_{x_i} \rightarrow 0,h_y \rightarrow 0}\frac{1}{4h_{x_i}h_{y_i}}\mathbb{E}[\mathbb{I}(x_i-h_{x_i}<X_i<x_i+h_{x_i},y_i-h_{y_i}<Y_i<y_i+h_{y_i})]
\end{split}
\end{equation} 

where $\mathbb{I}$ is the indicator function.

Given samples of the pair, the joint PDF can thus be be estimated as
\begin{equation}
\begin{split}
\widetilde P(x_i,y_i) & = \frac{1}{4n_ih_{x_i}h_{y_i}} \sum\limits_{j=1}^{n_i}[\mathbb{I}(x_i-h_{x_i}<x_{ij}<x_i+h_{x_i},y_i-h_{y_i}<y_{ij}<y_i+h_{y_i})] \\\
& = \frac{1}{4n_ih_{x_i}h_{y_i}} \sum\limits_{j=1}^{n_i}w_{h_{x_i}, h_{y_i}}\left((x_{i,j},y_{i,j}),(x_i,y_i) \right)
\end{split}
\end{equation} 
where 
\begin{equation}
w_{h_{x_i},h_{y_i}}((x_{i,j},y_{i,j}),(x_i,y_i)) =\mathbb{I}(x_i-h_{x_i}<x_{ij}<x_i+h_{x_i},\\y_i-h_{y_i}<y_{ij}<y_i+h_{y_i}) .
\end{equation} 

To enable the use of joint density $\widetilde P(x_i,y_i)$ by a CNN  we need to convert it to a 2D histogram. We thus obtain a 2D joint empirical PDF of size of $K\times K$ by setting:
\begin{equation}
EPDF_i (o,p)=  1/n_i  \sum\limits_{j = 1}^{n_i} \mathbb{I}\left(x_o \leq x_{ij} < x_{o+1} , y_p \leq y_{ij} < y_{p+1}\right) 
\end{equation}
Where $x_o (x_{o+1})$ is the minimal (maximal) x value for the $oth$ bin of x, $y_p (y_{p+1})$ is the minimal (maximal) y value for the $pth$ bin of y, and $1 \leq o,p \leq K$. In this paper we uniformly divided the observation ranges either in the original or in log space, though other ways that result in non uniform bins are also possible. We next normalize the $EPDF_i$ and obtain a Normalized $EPDF_i$ between $[0,1]$. 

Our representation of each pair by a $NEPDF$ converts the inputs to an image-like matrix of size of $K \times K$. This provides a number of advantages.  First, such representation preserves locality (unlike the original vector space \cite{yuan2019deep}).  Second, since the NEPDF matrix size is fixed, for each pair the training and test run time is fixed regardless of the number of observations. Finally, a very common problem in dealing with scientific dataset is the missing of observations, but the construction of $NEPDF$ can easily handle such cases without the need to add additional modifications to the observations.

When generating a train/test dataset for $NEPDF$, for each positive pair (x, y), the model will also learn a $NEPDF$ for (y, x) which will be assigned a label of -1. Note that the $NEPDF$, of (x, y), is the transpose matrix of (y, x). CNNs are known to learn localized features and thus will assign different values for such rotations. While this is a regarded as a problem for other applications \cite{shorten2019survey}, we  can take advantage of such property when learning the networks by using both the original and rotated versions of the $NEPDF$ with opposite labels.

\subsection{CNN classifier}
We next use the constructed $NEPDF$ to train a CNN based on VGGNet \cite{simonyan2014very}. The CNN consists of $q$ stacked convolutional layers of size $3 \times 3$, where $q$ is the power of 2, and interleaved maxpooling layer of size of $2 \times 2$.  The final layer is a logistic regression for binary label task and softmax classifier for multilabel task, so that the prediction is
\begin{equation}
\widetilde L_i = \mathop{\arg\max}_{l} CNN(NEPDF_i)
\end{equation}
with a crossentropy loss function
\begin{equation}
Loss = \sum\limits_{i = 1}^{N}\sum\limits_{l = 1}^{L}L_{i,l}log(y_{i,l})
\end{equation}
where, $L_{i,l} = 1 $ if sample $i$ belongs to $lth$ class, $y_{i,l}$ is the prediction probability that sample $i$ belongs to $lth$ class by CNN. Stochastic gradient descent (SGD) is used to train the model. Note that the size of $NEPDF$ is an important hyper-parameter for the CNN. As a result size can be optimized by cross validation. We provide an implementation of $NEPDF$ and CNN classifier in the github.\footnote{\url{https://github.com/xyvivian/causal_discovery_with_CNN.git}}. 

\subsection{Extension to multivariate case}
For multivariate cases, for example, drawing the pathway of RNA/DNA flows or determining the causal graphs between multiple cause and effect data, the CNN method can be used in two ways. The CNN method can be used in a general PC-algorithm \cite{spirtes2000causation} framework, and provide causal direction for pairwise data beyond V-structure cases. Another way of mining three-variable relationships is by construction of 3D $NEPDF$s.

\section{Experiments}
\label{experiments}

To test the usefulness of our CNN method, we conduct  series of experiments on both simulated and real data, and compare our algorithm to other state of the art methods including the baseline regression method (BivariateFit),  Randomized Causation Coefficient RCC \cite{lopez2015towards},  Neural Causation Coefficient NCC \cite{lopez2017discovering} ,Additive Noise Model ANM \cite{assaadANM}, Information Geometric
Causal Inference method IGCI \cite{janzing2012information},Regression Error based Causal Inference RECI \cite{bloebaum2018cause}, and Jarfo \cite{fonollosa2019conditional}.

\subsection{Numerical Simulations}

A Structure Equation Model \cite{pearl_2009} for an acyclic graph $\mathcal{G}(V,E)$ specifies for all $i \in V$ a function for the random variable $X^i = f_i(\textbf{PA}^i, N_i)$, where $\textbf{PA}^i$ denotes the parents of $X_i$ and $N_i, i \in V$ are jointly independent. In the bivariate case of $X$ and $Y$, the causal relationship can be expressed according to the function $Y = f(x) + n$  \cite{shimizu2011directlingam, inproceedings,  hoyer2009nonlinear}. Such additive function can be further extended to the time series model \cite{peters2013causal}: For each $i$ from a finite set of variables $V$ , let $(X^i_t)_{t \in \mathbb{N}}$ denotes the time series. Thus, the goal is to acquire a summary time graph, with the time series components as vertices and an edge between $X_i$ and $X_j$ if there is an edge from $X^i_{t-k}$ to $X^j_t$ for some $k$.

As a simplification to the time series model, we performed simulated experiments using three random variables $(X,Y,Z)$ with a skeleton network $X-Y-Z$. We selected the three most common causal structures for such variables in directed acyclic graphs (DAGS): $X \rightarrow Y \leftarrow  Z$ (V structure), $X \rightarrow  Y \rightarrow Z$ or $X \leftarrow Y \leftarrow Z$ (chain structure), and $X \leftarrow Y \rightarrow Z$ (Reverse V structure). For each of these causal structures we selected specific functions (see below) and repeated the simulation $10,000$ times with different random seeds. Then, we divided the generated causal pairs into training, validation and testing, and obtained the score using a five-fold cross validation analysis.

Since many of the methods we compared to formulate the problem as a binary classification task, we followed the approach presented in the ChaLearn's Challenge \cite{lopez2015towards} to evaluate performance.
Specifically, we trained two classifiers for causal/reverse-causal and dependent/independent predictions. Since these classifiers can lead to contradictory results (for example $y_{causal}$ and $y_{ind}$), the final prediction is computed as: $y_{pred} = y_{ind} \cdot (2 \cdot y_{causal} -1)$.

\subsubsection{V structure}
\label{V structure} We used the following non-Gaussian additive equations to construct the scenario where two variables share a common confounder with one time-lag. The equations were repeated $1,000$ times starting with a random seed to obtain 1000 values for our three variables (X, Y and Z).
\begin{align*}
 X_t &= \alpha  X_{t-1} +(1-\alpha)  X_{noise,t} \\
Z_t &= \alpha  Z_{t-1} +(1-\alpha)  Z_{noise,t} \\
Y_t &= \alpha  Y_{t-1} + \frac{\beta}{2}  ( X_{t-1}+  Y_{t-1}) + (1 - \beta - \alpha)  Y_{noise,t} \\
\end{align*}
where $X_{noise,t}, Z_{noise,t}$ and $Y_{noise,t}$ are generated from $\mathcal{N}(\mu,\sigma)$. At each step, the $\mu$ and $\sigma$ are sampled from the uniform distributions $\mathcal{U}[0,10]$ and $\mathcal{U}[0,50]$, respectively. We experimented with different settings of $\alpha$ and $\beta$ as we show in Table 1. Larger $\alpha$ and $\beta$ put more weight on the previous values whereas smaller ones put more weight on the influence of noise.

We used the simulation data to construct six 2D $NEPDF$s with ground truth labels: $(X, Y, 1), (Y, X, -1), (X, Z, 0), (Z, X, 0), (Z, Y, 1)$ and $(Y, Z, -1)$. Here the labels of $1$ and $-1$ stand for causal/ reverse-causal relationship, while $0$ means that the two variables are independent.  Table 1 shows average AUROC for one label versus the other two for all methods we tested. As can be seen, all methods perform less satisfactorily when the data is more noisy (lower $\alpha$ and $\beta$). However, when we increase the weight of past observation our CNN approach greatly outperforms all other methods in terms of AUC score. 

\begin{table*}[]
\label{Tab 1}
\centering
\caption{Experiment 1:  AUROC Score for V Structure Link Prediction. The performances of CNN and NCC become better when the $\alpha$ and $\beta$ increases, meaning less noisy data. However, for ANM, RCC, Jarfo, IGCI, RECI and BivariateFit, their AUC Scores have not shown much improvement.}
\begin{tabular*}{0.84\textwidth}{|l|llllllll|}
\hline
$\alpha$, $\beta$ & CNN & NCC  & ANM  & RCC  & Jarfo & IGCI & RECI & BivariateFit \\ \hline
0.1, 0.1          & \textbf{0.75} & 0.54 & 0.47 & 0.50 & 0.49  & 0.65 & 0.48 & 0.43         \\
0.1, 0.2          & \textbf{0.82} & 0.73 & 0.50 & 0.50 & 0.49  & 0.64 & 0.47 & 0.42         \\
0.1, 0.5          & \textbf{0.92} & 0.69 & 0.51 & 0.49 & 0.50  & 0.65 & 0.50 & 0.50         \\
0.2, 0.1          & \textbf{0.77} & 0.57 & 0.47 & 0.57 & 0.52  & 0.65 & 0.50 & 0.40         \\
0.2, 0.2          & \textbf{0.84} & 0.55 & 0.47 & 0.52 & 0.49  & 0.64 & 0.47 & 0.41         \\
0.5, 0.1          & \textbf{0.83} & 0.51 & 0.48 & 0.52 & 0.50  & 0.65 & 0.48 & 0.43         \\
0.5, 0.5          & \textbf{0.99} & 0.73 & 0.48 & 0.48 & 0.48  & 0.69 & 0.49 & 0.40         \\ \hline
\end{tabular*}
\end{table*}

\subsubsection{Chain Structure}
\label{Chain structure}
For the chain structure, $1,000$ time-step data pairs were simulated using the following non-linear equations with non-Gaussian noise:
\begin{align*}
X_t &= \alpha X_{t-1} +(1-\alpha) X_{noise,t} \\
Z_t &= \beta  Z_{t-1} + \gamma  (X_{t-1} - 1)^2 +(1- \gamma - \alpha)  Z_{noise,t}\\
Y_t &= \beta Z_{t-1} + \frac{\gamma}{2} (\cos( Z_{t-1}) + \sin(Z_{t-1}) + (1 -\beta - \gamma) Y_{noise,t}
\end{align*}
Again $X_{noise,t}, Z_{noise,t}$ and $Y_{noise,t}$ are generated from $\mathcal{N}(\mu,\sigma)^3$. At each step, $\mu \sim \mathcal{U}[0,10]$ and $\sigma \sim \mathcal{U}[0,50]$. The ground truth labels are $(X, Y, 0), (Y, X, 0), (X, Z, 1), (Z, X, -1), (Z, Y, 1)$ and $(Y, Z, -1)$. We again tested  different settings for $\alpha$, $\beta$ and $\gamma$. Table 2 shows the average AUROC for all methods. In all cases, CNN achieved an AUROC score above $0.9$, and it can successfully detect the independence between $X$ and $Y$ variable when $Z$ was the intermediate node. For this setting CNN achieved much better performance than all methods we compared to for all parameters we tested. 

\begin{table*}[]
\centering
\label{Tab 2}
\caption{Experiment 2:  AUROC Score for Chain Structure Link Prediction. CNN gives the highest AUC Scores across all different configurations of $\alpha$, $\beta$ and $\gamma$.}
\begin{tabular*}{0.885\textwidth}{|l|llllllll|}
\hline
$\alpha$, $\beta$, $\gamma$ & CNN & NCC  & ANM  & RCC  & Jarfo & IGCI & RECI & BivariateFit \\ \hline
0.1, 0.1, 0.1               & \textbf{0.93} & 0.79 & 0.52 & 0.49 & 0.50  & 0.53 & 0.49 & 0.48         \\
0.1, 0.2, 0.1               & \textbf{0.94} & 0.78 & 0.53 & 0.49 & 0.50  & 0.54 & 0.50 & 0.50         \\
0.1, 0.2, 0.5              & \textbf{0.99} & 0.79 & 0.52 & 0.50 & 0.51  & 0.53 & 0.50 & 0.56         \\
0.2, 0.1, 0.1              & \textbf{0.95} & 0.78 & 0.52 & 0.52 & 0.50  & 0.59 & 0.58 & 0.56         \\
0.2, 0.5, 0.5               & \textbf{0.97} & 0.80 & 0.59 & 0.50 & 0.50  & 0.53 & 0.50 & 0.50         \\
0.5, 0.1, 0.1               & \textbf{0.96} & 0.78 & 0.61 & 0.53 & 0.51  & 0.58 & 0.50 & 0.50         \\
0.5, 0.1, 0.2               & \textbf{0.98} & 0.79 & 0.60 & 0.50 & 0.50  & 0.60 & 0.49 & 0.50         \\
0.5, 0.5, 0.5               & \textbf{0.99} & 0.82 & 0.60 & 0.55 & 0.50  & 0.59 & 0.51 & 0.51         \\ \hline
\end{tabular*}
\end{table*}

\subsubsection{Reverse V-Structure}
\label{Reverse V structure}
The reverse-v structure involves two variables sharing one parent variable node. We generated $1,000$ time-step data pairs with ground truth labels $(X, Y, 1), (Y, X, -1), (X, Z, 0), (Z, X, 0), (Z, Y, 1)$ and $(Y, Z, -1)$. using the following equation:

\begin{align*}
Y_t &= \alpha Y_{t-1} +(1-\alpha) Y_{noise,t} \\
X_t &= \beta X_{t-1} + \gamma  Y_{t-1} +(1- \beta -\gamma)  X_{noise,t}\\
Z_t &= \beta  Z_{t-1} +  \gamma  Y_{t-1} + (1- \beta - \gamma)  Z_{noise,t}
\end{align*}

As before, we generated $X_{noise,t}, Z_{noise,t}$ and $Y_{noise,t}$ using $\mathcal{N}(\mu,\sigma)$, with $\mu \sim \mathcal{U}[0,10]$ and $\sigma \sim \mathcal{U}[0,50]$ at each step. Table 3 presents the average AUROC for each method.

\begin{table*}
\centering
\caption{Experiment 3: AUROC Score for Reverse V-structure Link Prediction. CNN and NCC have increment in AUROC score when the weight of past observations($\beta$ and $\gamma$) increases. }
\label{Tab 3}
\begin{tabular*}{0.885 \textwidth}{|l|llllllll|}
\hline
$\alpha$, $\beta$, $\gamma$ & CNN & NCC  & ANM  & RCC  & Jarfo & IGCI & RECI & BivariateFit \\ \hline
0.1, 0.1, 0.1               & \textbf{0.71} & 0.56 & 0.50 & 0.49 & 0.50  & 0.53 & 0.50 & 0.48         \\
0.1, 0.2, 0.1               & \textbf{0.84} & 0.63 & 0.53 & 0.48 & 0.49  & 0.53 & 0.52 & 0.50         \\
0.1, 0.2, 0.5               & \textbf{0.96} & 0.66 & 0.52 & 0.51 & 0.49  & 0.60 & 0.50 & 0.57         \\
0.2, 0.5, 0.1               & \textbf{0.98} & 0.48 & 0.59 & 0.51 & 0.48  & 0.58 & 0.51 & 0.57         \\
0.2, 0.5, 0.5               & \textbf{0.98} & 0.58 & 0.53 & 0.49 & 0.49  & 0.51 & 0.50 & 0.49         \\
0.5, 0.1, 0.1               & \textbf{0.88} & 0.67 & 0.50 & 0.51 & 0.49  & 0.51 & 0.50 & 0.50         \\
0.5, 0.2, 0.1               & \textbf{0.96} & 0.68 & 0.48 & 0.50 & 0.50  & 0.50 & 0.50 & 0.50         \\
0.5, 0.5, 0.5               & \textbf{0.99} & 0.73 & 0.50 & 0.53 & 0.49  & 0.52 & 0.50 & 0.51         \\ \hline
\end{tabular*}
\end{table*}

\subsection{Tübingen dataset} The Tübingen dataset consists of $108$ heterogeneous, hand-crafted $X-Y$ pairs from real world causal discoveries, including physical phenomenons, social and economical effects, biological observations, and etc \cite{JMLR:v17:14-518}. Since this dataset is fairly small, it is not possible to generate sufficient $NEPDF$s for CNN to train only from this data. Instead, we followed \cite{lopez2017discovering} and constructed a larger dataset by adding heteroscedastic additive noise to obtain additional pairs. We used this to construct $15,000$ synthetic samples, where each sample $S_i$ included $m_i \sim \mathcal{U}[100, 1000]$ pairs of $\{x_{ij}, y_{ij}\}_{j =1}^{m_i}$. 

A similar $X-Y$ pair generation was used in our experiment \cite{lopez2017discovering}, where each of the cause $x_{ij}$ was drawm from a $k_i$-mixture Gaussian distribution, with each Gaussain sampled with mean $\mu_i$ and variance $\sigma_i^2$($k_i \sim$ Integer($\mathcal{U}[1, 5]$), $\mu_i, \sigma_i \sim \mathcal{U}[0, 5]$). $y_{ij}$ is generated from $y_{ij} = f_i(x_{ij}) + v_{ij}e_{ij}$, where $f_i$ is a 5-knot cubic Hermite spline function with support in $[\min(\{x_{ij}\}_{j=1}^{m_i}) - \text{std}(\{x_{ij}\}_{j=1}^{m_i}),  \max(\{x_{ij}\}_{j=1}^{m_i})+ \text{std}(\{x_{ij}\}_{j=1}^{m_i})]$. The noise terms $e_{ij}$ are sampled from $\mathcal{N}(0, v_i)$, where $v_i \sim \mathcal{U}(0, 5) $.

From this we constructed  $NEPDF$ with size of $16 \times 16$ for both $(x_{ij}, y_{ij})$ (label $-1$) and $(y_{ij}, x_{ij})$ (label $1$). For CNN training, we applied five-fold cross validation and achieved an average test accuracy of $0.913$ on the synthetic causal pairs. 

In the directional classification task for Tübingen dataset, we calculated a weighted accuracy score to avoid excess similarity in the testing data, as suggested by \cite{JMLR:v17:14-518}. The  classification accuracy of our CNN approach was $0.784$, which is comparable to performance reported for state-of-the-art methods NCC ($0.79$) \cite{lopez2017discovering} and RCC($0.75$) \cite{lopez2015towards}.

\subsection{Gene relationship inference }
\label{pure pairwise causality prediction}
While causal inference is challenging in all scientific areas, following decades of research biologist have been able to map several causal interactions between genes using knockout experiments. Current knowledge about such interactions is often encoded in the form of genetic pathways in which directed edges connect causal genes\cite{croft2014reactome}. We thus tested our method using data from the Kyoto Encyclopedia of Genes and Genomes (KEGG) pathway database \cite{kanehisa2017kegg} which compiles a set of experimentally validated gene pathways graphs where each node represents one gene and each edge represents directed or undirected interaction.  For our analysis we focused on the directed gene pairs for the positive labels. As for the observation data, we used the Encyclopedia of DNA Elements (ENCODE) \cite{yue2014comparative} gene expression dataset that contains 249 samples, each of which has more than forty thousand genes. 

We compared our method with the same set of prior methods we used for the simulated data. Given a known gene pair $gene1 \rightarrow gene2$, a $NEPDF$ of $(gene1, gene2)$ with label 1 and a $NEPDF$ of $(gene2, gene1)$ with label -1 were generated. Table 4 shows the comparison result with three-fold cross validation. For RCC, several combinations of hyperparameters were used, and we reported the series of hyperparameters with the best performance. As can be seen, our CNN method outperformed all prior methods for gene causality prediction. Interestingly, unlike for the simulation data we observed that only the supervised learning methods were able to perform well on this task whereas all the unsupervised methods, including ANM, RECI, IGCI, BivariateFIT and CDS do not perform well. This may indicate that for real world data (including gene regulation network inference) the strong assumptions used by unsupervised methods about the statistical model of dependency may lead to inaccurate results.

\begin{table*}
      \caption{The AUROC Score for gene edge direction prediction with KEGG pathway database.}
  \centering
  \label{Tab 4}
  \begin{tabular*}{0.932\textwidth}{|l|llllllll|}
  \hline
    Task     & ANM & RECI & IGCI & BivariateFit & RCC  & NCC & CDS & CNN  \\\hline
    Causality task & 0.498 & 0.503 & 0.509 & 0.458 & 0.620 & 0.699 & 0.543   & \textbf{0.743}   \\\hline
  \end{tabular*}
\end{table*}

\begin{table*}
     \caption{Test Bidirection AUROC Score for Chalearn's Challenge Data.}
  \centering
  \label{Tab 5}
  \begin{tabular*}{0.962 \textwidth}{|l|lllllll|}
  \hline
    Task     & Jarfo & RECI & IGCI & BivariateFit & RCC  & NCC & CNN  \\\hline
   Bidirectional AUROC Score & \textbf{0.82} & 0.49 & 0.51 & 0.59 & 0.74 & 0.55 & 0.74  \\\hline
  \end{tabular*}
\end{table*}

\subsection{ChaLearn's Challenge Data(Kaggle Competition)}

The Kaggle causal-effect pairs competition collected data from different scientific areas \cite{guyoncause}. It provided more than $15,000$ variable pairs, each of which has thousands of observation values and a target label vector. As with our experiments, there were labels of $1$ and $-1$ meaning the causal and anti-causal relationships, $0$ indicating no causal relationship. The competition also provided a test dataset of $4,050$ pairs with observation values but without target labels. On the test data, our single neural network model achieved a mean test test bidirectional AUC of $0.74$, which is comparable with the score obtained by RCC  ($0.74$) \cite{lopez2015towards}. Note that to run our method we only need a single CNN network trained on NEPDF inputs. In contrast, RCC uses a large engineered feature vector and an ensemble learning strategy based on tens of classifiers. While $0.74$ is less than the winning score ($0.82$) of the winning method, Jarfo. However, Jarfo was heavily tailored to the data and so may not generalize to other causality prediction tasks. Table 5 shows the comparison result we compare with other supervised and unsupervised baselines.

\subsection{Stock sector classification}
\label{arbitrary defined relationship prediction}
While our main focus is on causal inference, our method can be trained on any type of dependency label a user supplies. To illustrate a potential use for such training we used our method to predict relationships between pairs of stocks. Specifically, we trained a model to predict if two stocks are from the same sector based on their price information. For this we downloaded data from Yahoo Finance covering the years between 2014 and 2019, and containing both price and sector information for all stocks in the S\&P 500 index. This dataset contained $1259$ observed prices for each stock. Data was processed to compute the daily changing percentage of stock prices which was used as an input to construct $NEPDFs$.We divided the stock into five equal size folds for cross-validation analysis. For training, we assigned a label of 1 if stocks $x_i$ and $x_j$ belong to the same sector and $0$ otherwise. 

\begin{figure}
  \centering
  \includegraphics[width=7.5cm]{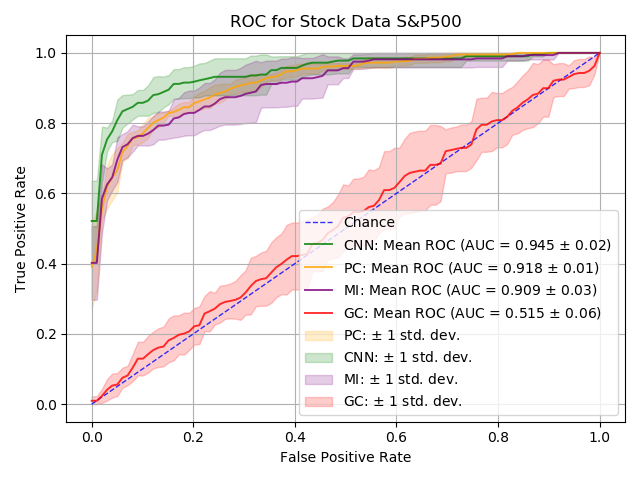}
  \caption{Results for CNN, PC, MI, and GC for stock sector prediction. While all CNN, PC and MI have shown increases in AUROC score than simple random guess and GC, CNN is also able to obtain a larger AUROC.}
  \label{Figure 3}
\end{figure}

Most of the methods we compared to above focus on predicting the {\em direction} of an interaction and cannot be used to predict new interactions. Thus, unlike our method they cannot be used to infer interactions between stocks. We thus compared our results with the most popular statistical measure, Pearson correlation(PC) and mutual information(MI), as well as Granger's causality(GC), which is a popular method to determine the relationships in time-series data. For Granger's causality, we use the p-values as the final prediction. Figure~\ref{Figure 3} presents the comparison between the methods we tested. As can be seen, our method achieved a  higher AUROC when compared to the other two methods. While PC and MI performed well, CNN still improved on MI for this task as well (AUROC of $0.945$ vs. MI($0.908$) and PC($0.918$)). 

\section{Conclusion and Future Work}
In this paper, we presented an efficient method to learn casual relationships that does not rely on an explicit statistical model. Our method converts general pair-wise observation data to normalized empirical probability density distribution ($NEPDF$) and uses this representation to train a convolutional neural network to predict causal or other relationships between input samples. The use of NEPDFs enables our method to utilize methods that work very well for image classification problems for the task of causal inference. Experiments on simulated and real data demonstrates that the method is general and works well on a diverse set of input data types. 

There are several ways in which we can further improve the methods and its results. First, improving the $NEPDF$ construction using other density estimation methods can lead to better representation and results. Second, we would like to extend pairwise inference that is currently supported to learn interactions and causality between a larger set of variables. Finally, we would like to use the method and representation to infer patterns that are indicative of causal relationships and to interpret these to better understand the underlying mechanisms that lead to the causal relationship observed. 

\section*{Broader Impact}

Here we present a general pair-wise variable relationship inference method based on probability estimation and deep learning technologies without relying on an explicit statistical model. Such method would benefit a range of scientific fields that are interested in pair-wise variable relationship exploration while lack known models, for example, gene regulation in biology, stock relationship in economy, climate changes impact among different regions and so on.

\bibliography{main} 
\bibliographystyle{plain}

\end{document}